\documentclass[11pt]{article}

\usepackage[utf8]{inputenc}
\usepackage[T1]{fontenc}
\usepackage[margin=1in]{geometry}
\usepackage{amsmath,amssymb}
\usepackage{booktabs}
\usepackage{xcolor}
\usepackage{enumitem}
\usepackage{listings}
\usepackage[hidelinks]{hyperref}
\usepackage{titlesec}
\usepackage{tikz}
\usepackage{tcolorbox}
\usepackage{xcolor}
\usepackage{soul}
\usepackage{adjustbox}

\usepackage{booktabs}
\usepackage{tabularx}
\usepackage{array}

\usepackage[table]{xcolor}

\definecolor{hmzero}{HTML}{FFFFFF}     
\definecolor{hmone}{HTML}{EAEAEA}      
\definecolor{hmtwo}{HTML}{FFF9C4}      
\definecolor{hmthree}{HTML}{E3F2FD}    
\definecolor{hmfour}{HTML}{C8E6C9}     

\newcommand{\scZ}{\cellcolor{hmzero}0}
\newcommand{\scO}{\cellcolor{hmone}1}
\newcommand{\scT}{\cellcolor{hmtwo}2}
\newcommand{\scH}{\cellcolor{hmthree}3}
\newcommand{\scF}{\cellcolor{hmfour}\textbf{4}}

\usetikzlibrary{positioning, calc, fit, backgrounds, arrows, shapes.geometric}

\titleformat{\section}{\normalfont\large\bfseries}{\thesection}{0.7em}{}
\titleformat{\subsection}{\normalfont\normalsize\bfseries}{\thesubsection}{0.6em}{}

\definecolor{codebg}{rgb}{0.97,0.97,0.95}
\title{Networked Intelligence: Active Shared Context Graphs for Human-AI Team Science}

\author{
Sutanay Choudhury\thanks{Corresponding author: \texttt{Sutanay.Choudhury@pnnl.gov}},
Jeffrey J. Czajka, Lummy M. O. Monteiro, Erin Bredeweg, \\ Jason McDermott, Katherine Wolf, 
Alex Beliaev, Josh Elmore, Paul Piehowski, \\ Kylee Tate, 
Yuqian Gao, Aivett Bilbao, Kelly Stratton, Scott Baker,\\ Jaydeep P. Bardhan,
Kristin Burnum Johnson, Chris Oehmen, Robert Rallo\\[0.5em]
\small Pacific Northwest National Laboratory\\
\small Richland, WA, USA
}

\begin{document}
\maketitle

\begin{abstract}
\noindent
Most AI-for-science systems focus on scaling a single reasoning process by using better models, larger context windows, long-horizon agentic execution, or digital co-scientists working with one principal user. However, challenging scientific problems are rarely solved by one reasoner alone. They are solved by teams whose members carry different priors, experimental background, tacit knowledge, and domain-trained intuitions. The open problem is therefore not only how to scale models, but how to develop \emph{networked intelligence}, scaling the connections between humans and AI systems so that a result or hypothesis produced in one context reaches another person, agent, instrument or robot that can act on it.  We introduce \emph{Mycelium}, an active shared workspace that automatically connects researchers and AI agents. As human users and agents work, the system captures important observations and hypotheses, tracks how they relate to the team’s evolving knowledge model, and routes them to the person or agent whose next decision they can inform. We evaluate Mycelium through a real-world scientific discovery use case: a biological multi-omics campaign where shared context turned a local analytical finding into a cross-expert mechanistic constraint and ultimately into an experimental design. Finally, we describe networked intelligence as sparse conditional computation over distributed scientific contexts. This framework establishes when a scaled standalone agent is sufficient, and when isolated data and specialized expertise make a networked approach essential.
\end{abstract}

\section{Introduction}

The next era of scientific discovery requires bridging two defining dimensions: the distributed expertise of team science and the autonomous execution of AI agents operating at machine speed. Most consequential discoveries now require expertise that exceeds what any individual can command, a mode the National Research Council formalized as \emph{team science} \cite{nrc2015teamscience}. AI systems have already begun to automate core scientific functions, from protein-structure prediction \cite{jumper2021highly} to experimental chemistry planning and execution \cite{boiko2023coscientist, pmlr-v235-sprueill24a, bran2024chemcrow, smith2026using}.  Driven by the convergence of these two trends, AI-accelerated team science is becoming a national-scale infrastructure priority, with initiatives like the DOE's Genesis Mission envisioning discovery platforms that connect isolated tools, instruments, and high-performance computing resources into shared resource networks \cite{ushizima2026report, doegenesis, miller2023integrated}. Advances in scientific agents, tools, and instruments are occurring at a faster pace than the knowledge architecture required to coordinate them.

The limiting problem lies in moving scientific context reliably among scientists, agents, and instruments while the team works on a scientific discovery problem. Existing AI co-scientists and autonomous laboratory workflows largely remain bounded by a single scientist's AI session or an orchestrated agent team \cite{autogen2023, tran2025collab}, even as the overarching scientific mission is distributed across specialists and still advances through slow, human-mediated exchange \cite{swanson2025virtuallab, gao2024biomedical}. As agents take on longer and more autonomous research tasks \cite{gridach2025survey}, coordination becomes an architectural bottleneck that model scaling alone does not remove: a model in one scientist's session can summarize local work, but it has no runtime mechanism for deciding when a finding should change another scientist's analysis, maintaining the team's evolving scientific hypotheses as persistent shared state, or guaranteeing that a propagated claim carries the provenance required for reuse. Agent-communication protocols such as the Model Context Protocol and Agent-to-Agent Protocol standardize how tools and agents discover or route tasks \cite{mcp, a2a, ehtesham2025survey}, while emerging memory layers persist an agent's own state across sessions \cite{mem0, ump, memorysurvey2025}. What remains as a key technical gap is a runtime architecture for \emph{networked intelligence} that allows scientific context to move across humans, agents, and instruments while remaining attributable, contestable, and actionable.

We introduce \emph{Mycelium}, a runtime architecture for networked human--AI scientific discovery built around an \emph{active context graph} (ACG). The graph represents shared scientific context as provenance-aware project state with typed entries for observations, interpretations, hypotheses, findings, open questions, recommendations, and experiment proposals, each carrying its derivation history. 

Mycelium makes three operations explicit. It routes context across users when one participant's work is relevant to another. It preserves the team's evolving knowledge model across sessions and actors.  And it keeps every routed claim tied to the evidence and reasoning that produced it. Human users and AI agents can read from and write to this shared graph through chat interfaces. As human and agentic users modify the graph through their work, Mycelium periodically discovers new connections among entries, updates the network state, and surfaces goal-relevant context to the users or agents whose next decisions it can inform.

We show that scaling the network, rather than just the model, alters the trajectory of an active scientific investigation. To isolate the value of this networked architecture, we compare Mycelium with a baseline.  The baseline is a standalone agent with access to the same data, maximum resources and full autonomy to reason across domains.  We evaluate Mycelium in a multi-omics microbial phenotyping campaign \cite{doe2026opal} in which three researchers with complementary expertise in proteomics, regulatory biology, and genetics worked asynchronously through their own AI chatbot interfaces alongside autonomous Mycelium investigations. Although the system is domain-neutral by construction, this campaign provides a preliminary empirical test and evaluation. 

The document is structured as follows. We first define the requirements for active shared-context networks and present the Mycelium architecture (Section 2). In Section 3, we detail this empirical test, demonstrating how routed context changed scientific interpretation and experimental action compared to the standalone-agent baseline. Finally, we formalize networked intelligence as sparse conditional computation, clarifying the strict boundaries where independent expertise makes a network irreducible to a single scaled model (Section 4).

\section{Methods}

\begin{figure}[h]
\centering
\includegraphics[width=1\linewidth]{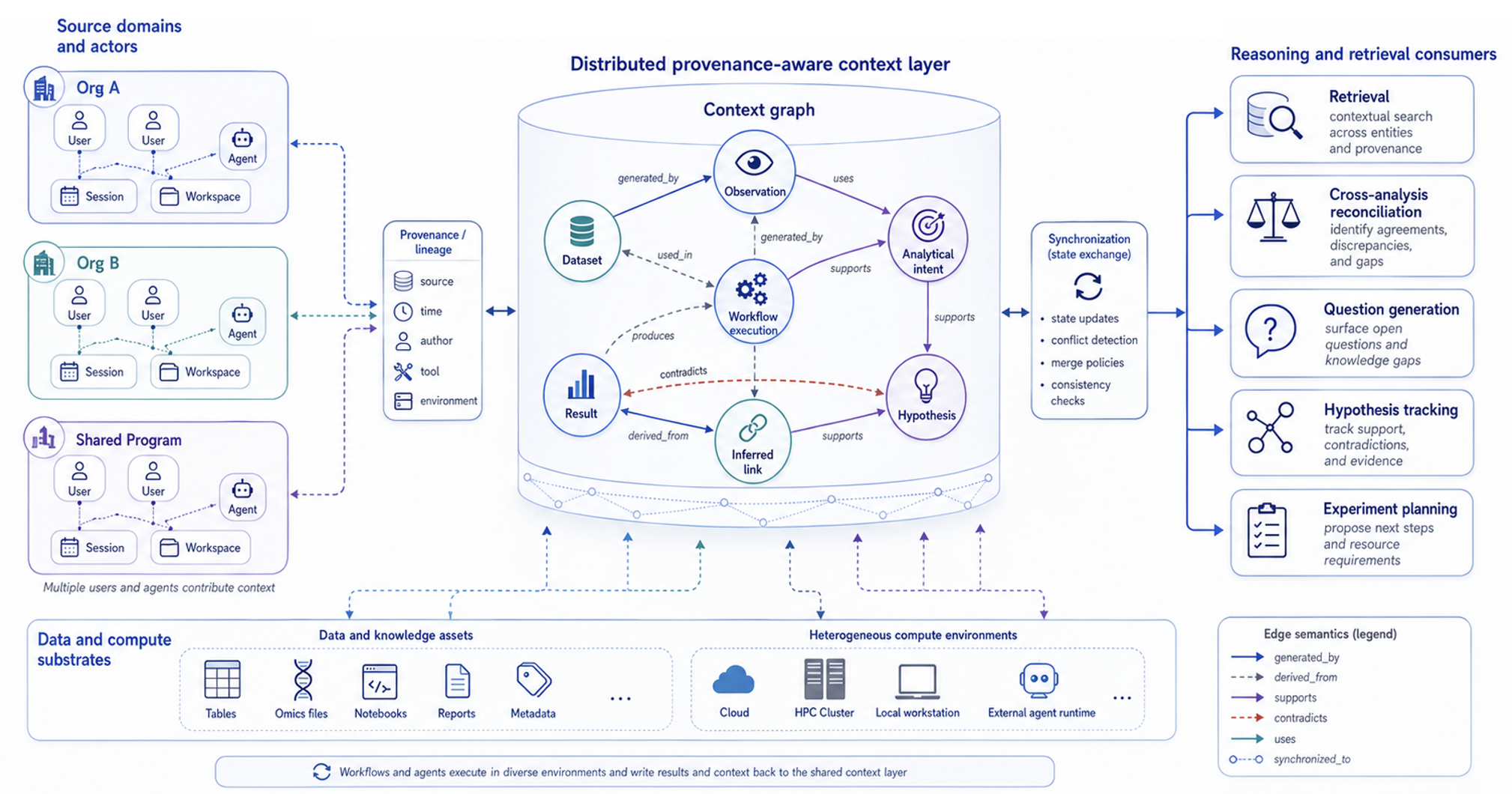}
\caption{\textbf{The Mycelium runtime architecture.} The system coordinates distributed scientific workflows across four distinct domains: \textbf{Source domains and actors (Left):} Researchers and AI agents operate within shared workspaces, reading and writing typed entries to the shared graph. \textbf{Active context graph (Center):} The core routing layer managing persistent project state. It maps provenance-aware relations (e.g., \textit{generated\_by}, \textit{supports}, \textit{contradicts}) among datasets, autonomous workflow executions, and evolving hypotheses. \textbf{Data and compute resources (Bottom):} Connects the runtime to scientific assets (e.g., omics files, notebooks) and execution environments (ranging from a virtual machine in cloud to HPC clusters). \textbf{Reasoning consumers (Right):} Specialized tools and co-scientists that read the synchronized graph to drive cross-analysis reconciliation, track hypotheses, and propose experimental designs.}
\label{fig:architecture}
\end{figure}

Mycelium implements a runtime architecture for networked human-AI scientific discovery (Fig. \ref{fig:architecture}). The architecture is built on three distinct pillars: (1) an active context graph that holds the shared scientific state and enforces provenance (Section 2.1); (2) a participant-facing runtime and API that connects researchers' standard AI chatbot clients to the server-side execution environment (Section 2.2); and (3) a set of shared-context protocols, the agentic primitives that autonomously inject, maintain, and route state across the distributed team (Section 2.3).

\subsection{Active context graph}

The key enabler of networked intelligence is an \emph{active context graph}, representing the shared project state as a directed graph, $G=(V, E)$. The nodes $V$ represent typed scientific entries, each carrying an epistemic role (e.g., evidence, reasoning, action) as defined by an ontology (Table \ref{tab:scp_entry_types}). The edges $E$ map cross-session provenance, linking every claim to its origin. \textbf{Lineage edges} track system operations (\texttt{derived\_from} for provenance, \texttt{generated\_by}/\texttt{produces} for tool execution, and \texttt{uses}/\texttt{used\_in} for dependencies), while \textbf{epistemic edges} map scientific belief states (\texttt{supports}, \texttt{contradicts}) across disparate user threads.  While the system enforces an entry vocabulary (Table ~\ref{tab:scp_entry_types}), conceptual visualizations of the graph (Figure \ref{fig:architecture}) group these primitives in terms of logical functions. For example, an analytical \emph{Result} is stored formally as a \texttt{finding} entry; an \emph{Analytical intent} represents a collection of queued-work and \texttt{open\_question} entries; and an \emph{Inferred link} is recorded as an \texttt{interpretation} bound by \texttt{derived\_from} provenance.

\begin{table}[h]
\centering
\caption{Closed entry-type vocabulary used by the active context graph.}
\label{tab:scp_entry_types}
\begin{tabular}{p{0.20\linewidth} p{0.16\linewidth} p{0.52\linewidth}}
\hline
\textbf{Entry Type} & \textbf{Role} & \textbf{Description} \\
\hline

\texttt{dataset} &
Evidence &
A registered data file or measurement table. \\

\texttt{observation} &
Evidence &
A raw or auto-captured analytical result. \\

\texttt{interpretation} &
Reasoning &
A participant's reading of an observation. \\

\texttt{hypothesis} &
Reasoning &
A testable claim. \\

\texttt{finding} &
Knowledge &
An established result. \\

\texttt{open\_question} &
Coordination &
A blocking question or knowledge gap. \\

\texttt{recommendation} &
Action &
A proposed next action. \\

\texttt{experiment\_proposal} &
Action &
A structured experiment with required resources. \\

\hline
\end{tabular}

\end{table}

We term this graph \emph{active} because the Mycelium runtime continuously evolves and evaluates the state, unlike a passive graph database. Specifically, the runtime executes three operations over this graph that are unavailable to standalone-agent sessions. \emph{Cross-user state routing} dynamically moves relevant context from an originating participant or process to the participant whose analysis it impacts, without requiring explicit requests. \emph{Evolving hypothesis state} maintains the team's claims alongside supporting and contradicting evidence across independent sessions. \emph{Provenance-bounded propagation} ensures that as claims are routed, they retain the lineage, tool chain, and attribution required for evaluation and reuse. Consequently, every downstream interpretation or proposed experiment is formally grounded in prior graph context, ensuring the entire collaborative lineage remains physically auditable.

\subsection{Runtime infrastructure for autonomous analytical execution}

Users interface with Mycelium through a standard AI chat client connected via the Model Context Protocol (MCP) (Fig. \ref{sfig:ui}). However, unlike standard chat models that lose context when a window closes, this runtime binds the client to the active context graph through a suite of stateful agentic tools. This allows the user's interface to seamlessly query the shared project state, dispatch autonomous workflows, and formally commit new findings.

\begin{figure}[h]\centering\includegraphics[width=0.75\linewidth]{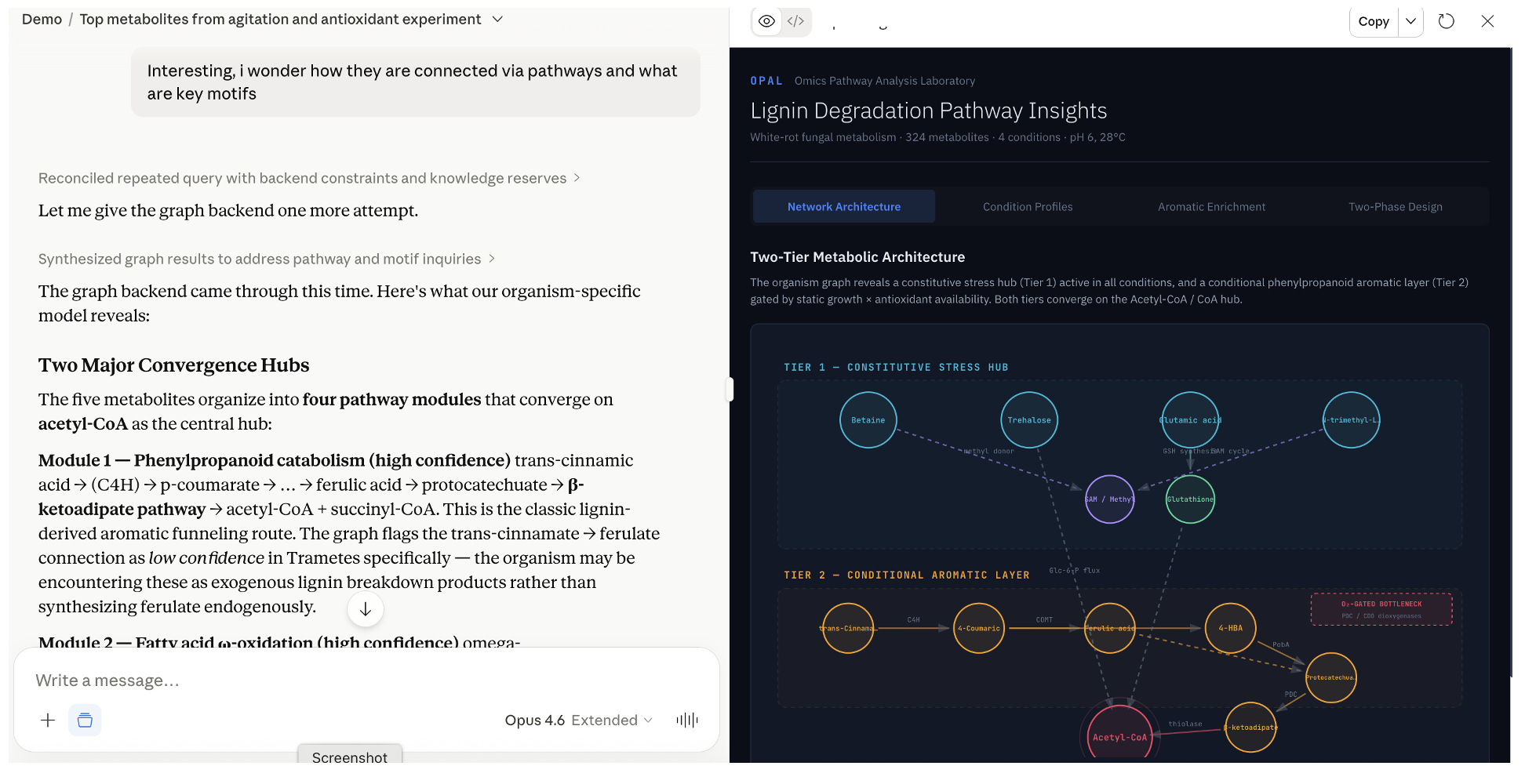}
\caption{\textbf{Participant-facing runtime.} Researchers interact with the Mycelium network using a standard, familiar AI chat interface. Rather than functioning as an isolated chatbot, MCP connects this chat window directly to the team's shared active context graph. This allows a user to query the entire project's history, dispatch complex autonomous workflows, and review results, all without leaving a simple conversational interface.}
\label{sfig:ui}
\label{sfig:ui}\end{figure}

To execute complex analytical workflows, the server-side runtime relies predominantly on dynamic, on-the-fly code generation within a sandboxed Python environment (Fig. \ref{sfig:dags}). The runtime integrates project-specific tools for standard tasks (e.g., data ingestion, PCA, differential abundance, and metabolic-pathway queries), preferentially leveraging validated library functions to enforce scientific reproducibility. It generates novel execution graphs only when a workflow demands analytical motifs that are unavailable. This autonomous execution includes automated error recovery. When a programmatic step fails due to inconsistent data or a runtime exception, the system reads the error trace, generates a corrective script, and resumes execution. Because the system is strictly schema-aware, it programmatically verifies all data structures prior to any interaction and emits a fully executable Jupyter notebook of its trace upon completion. This ensures that every autonomous discovery remains fully reproducible and auditable by a human expert.

\begin{figure}[h]\centering\includegraphics[width=0.8\linewidth]{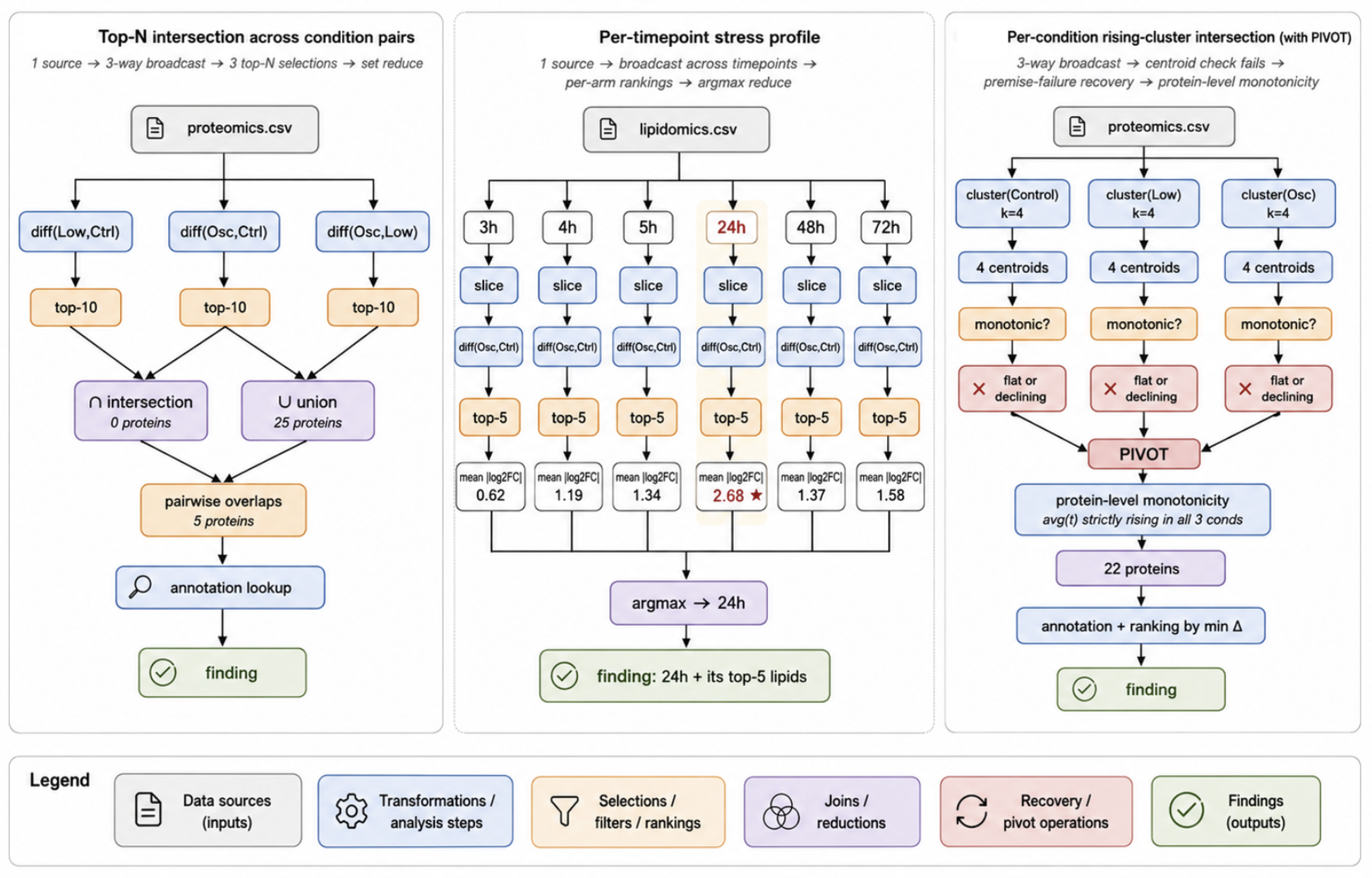}
\caption{\textbf{Autonomous workflow execution.} The Mycelium runtime leverages dynamic code generation to construct multi-step analytical pipelines on the fly for both interactive and autonomous analysis. The figure above shows illustrative computational dataflow graphs. The execution engine supports automated fault tolerance, allowing the system to recover from runtime errors and successfully output formalized findings to the shared project state.}
\label{sfig:dags}
\label{sfig:dags}\end{figure}

\subsection{Shared-context protocols}

Four operational primitives define how scientific knowledge enters, moves through, and is reused across the Mycelium network. These primitives make the active context graph more than a shared memory store. They specify how claims become auditable state, how autonomous work remains bounded, how relevant context reaches the right recipient, and how synthesized claims remain tied to evidence.

\textbf{Provenance (Logging).} The runtime enforces strict provenance by converting analytical executions into persistent, queryable state nodes. When a human or agent executes a task, the system automatically captures the result as an immutable observation bound exactly to its parameter space, tool chain, and dataset references.

\textbf{Bounded autonomy (Proactive Execution).} The runtime continuously monitors for emerging knowledge gaps and launches autonomous exploratory analyses between active user sessions (Fig. \ref{sfig:workload} in Supplementary Methods). When a structural defect or missing link is detected in the data, the agent autonomously plans and executes a corrective task. The rectified result is then persisted to the active context graph, cleanly surfacing for the participant as an actionable finding upon their next session.

\textbf{State-routing (Pollination).} The defining capability of the network is the asynchronous routing of emerging context to relevant participants (see Fig. \ref{fig:propagation} in Supplementary Methods). To achieve this without overwhelming users, the system evaluates new candidate linkages against localized belief states to score their epistemic utility, determining if a new observation supports, refines, or introduces tension to a specific researcher's ongoing analysis. Algorithmically, this scoring-and-updating operation across neighboring threads acts as a message-passing step \cite{pearl2022fusion, pmlr-v70-gilmer17a}. However, unlike standard graph algorithms that average signals into a single joint distribution, this architecture deliberately preserves disagreement.  Belief states remain isolated to their respective threads, ensuring that scientific contradictions across distinct analytical perspectives remain distinct objects of attention rather than being silently merged.


\textbf{Grounding (Persistence).} The runtime enforces strict validation at the graph-write layer for all scientific entries (e.g., \texttt{hypothesis}, \texttt{observation}, \texttt{finding}, \texttt{experiment\_proposal}). Every entry committed to the active context graph must be explicitly grounded in existing graph nodes, such as registered datasets, verified tool outputs, or provenance-bearing parent entries. The grounding requirement keeps synthesized claims tied to their empirical lineage.  It produces more specific scientific artifacts and prevents ungrounded, unstructured evidence from accumulating within the project state.

\section{Results}

This work frames the test of networked intelligence as a domain-neutral coordination problem, exploring whether shared context routed via an active context graph can change scientific interpretation and experimental action across distributed users. The evaluation used specific, testable outcomes.  Using the proposed active graph approach, local results are  stored and propagated to reach the scientis or agent context where it becomes an actionable constraint, shifts an interpretation, or alters an experimental decision.

As a case study, we deployed Mycelium in a microbial multi-omics campaign investigating secreted gluconate accumulation \cite{bentley2020engineering} in \emph{Pseudomonas putida}, an industrial bioproduction chassis \cite{nikel2018pseudomonas}. The objective was to extract actionable insights across four data modalities to inform the next round of strain and medium designs for PNNL's Anaerobic Microbial Phenotyping Platform (AMP2) \cite{ushizima2026report, smith2026using}.  In the following subsections, we first detail the distributed scientific setup (Section 3.1). Next, we analyze the empirical impact of active context routing on the team's scientific findings (Section 3.2). Finally, we benchmark this networked execution against standalone-agent baselines to quantitatively and qualitatively isolate the value of network scaling over model scaling (Section 3.3).

\subsection{Scientific campaign setup}
\label{sec:Scientific case-study setup}

\begin{figure}[h]
\centering
\includegraphics[width=1\linewidth]{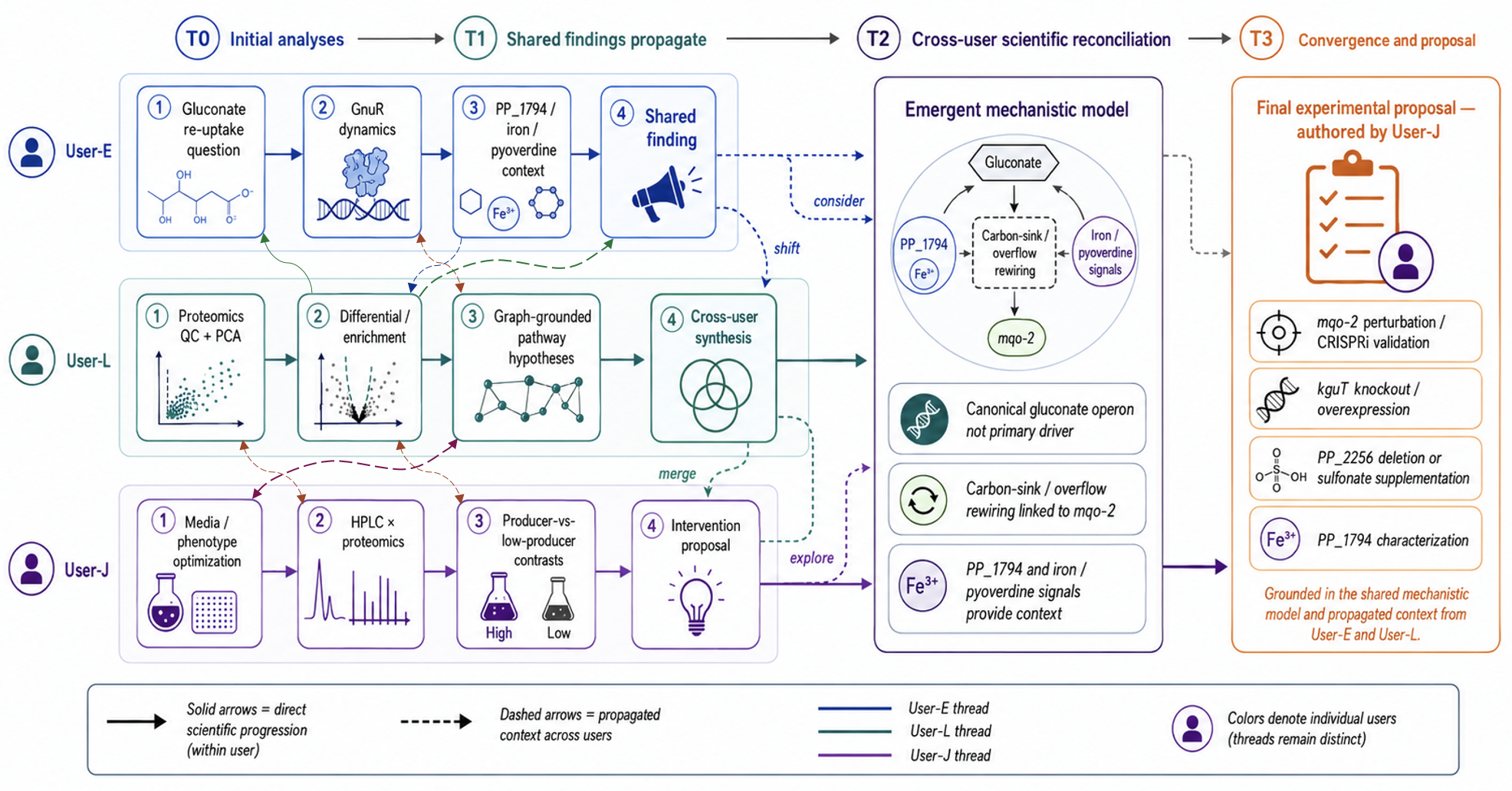}

\caption{\textbf{Context routing enables emergent collaborative discovery.} The campaign is partitioned into three expert threads (rows): regulatory reasoning (User-E), proteomic/pathway analysis (User-L), and phenotype-guided design (User-J). Solid arrows represent intra-thread reasoning; dashed arrows represent context routed by Mycelium across participants. }

\label{fig:convergence}
\end{figure}

Over one week, three domain experts analyzed the campaign asynchronously through independent chatbot interfaces (Fig. \ref{fig:convergence}). User-L performed proteomics and systems analysis on a differential-abundance dataset of 4,495 proteins across 62 samples. User-E provided regulatory and iron-biology reasoning, and User-J contributed phenotype interpretation and experimental design, including high-performance liquid chromatography (HPLC) assays, among other analyses detailed below. The campaign spans four data modalities across four engineered strains (Supplementary Table~\ref{stab:dataset}). All researchers worked through Claude chat-based Mycelium client (as shown in Fig. \ref{sfig:ui}). All client-and server-side Mycelium operations used Claude Opus 4.8.

\subsection{Enabling team coordination via shared context propagation}
\label{sec:team-convergence}

Team science depends on two coordination challenges: reconciling distributed findings across isolated domains, and producing synergistic outcomes that a single researcher cannot achieve alone. The following results show how Mycelium addressed both.

\textbf{Reconciling multiple sources of evidence:} The first instance of connecting distributed findings showed how the network converted an isolated negative result into corroborating evidence. User-E (focusing on biological regulation) noted that the expected genetic pathways for processing the target chemical (canonical gluconate-uptake and Entner-Doudoroff pathway) were not turning on as anticipated \cite{bentley2020engineering}. When routed into User-L's thread, this regulatory anomaly recontextualized an independent finding: the corresponding proteins were also absent from the top proteomic signals. Combined via a \texttt{derived\_from} edge, these isolated observations successfully reconciled a 12-h regulatory differential with 24-h proteomics. Fig.~\ref{fig:convergence} shows this interaction from User-E's regulatory thread (top row) through User-L's proteomic analysis to the emergent model (T2).

\textbf{Integrating cross-domain evidence and team constraints:} The second instance of connecting distributed findings showed how cross-domain evidence shifted the analytical approach. User-J (focusing on physical outcomes) uploaded chemical measurements (HPLC) that distinguished successful, high-producing bacterial strains from low-producing ones. When this constraint was routed into User-L's workspace, it reframed the proteomic analysis. It shifted User-L's synthesis from a standard protein-ranking problem to an intervention problem, requiring the identification of testable molecular mechanisms to explain the measurable extracellular accumulation. Fig.~\ref{fig:convergence} shows where evidence from User-J's phenotype thread (bottom row) is routed into User-L's analysis.

\textbf{Steering the team to convergence:} In the campaign, continuous routing assembled a mechanistic model that no single participant held. User-E supplied the regulatory contradiction; User-L provided proteomic evidence for carbon-sink rewiring; and User-J imposed the phenotype constraint. The resulting model focused attention on gluconate retention and re-assimilation rather than synthesis. Shared context also also identified an association between producing strains and elevated sulfonate-scavenging and sulfur-metabolism proteins \cite{kertesz2000riding}. The network also pruned false leads; For example, cross-checking gene identifiers against growth-state signatures redirected a control variable to a characterization marker (\emph{Emergent mechanistic model} panel,  Fig.~\ref{fig:convergence}).

\begin{figure}[!hb]
\centering
\includegraphics[width=0.7\linewidth]{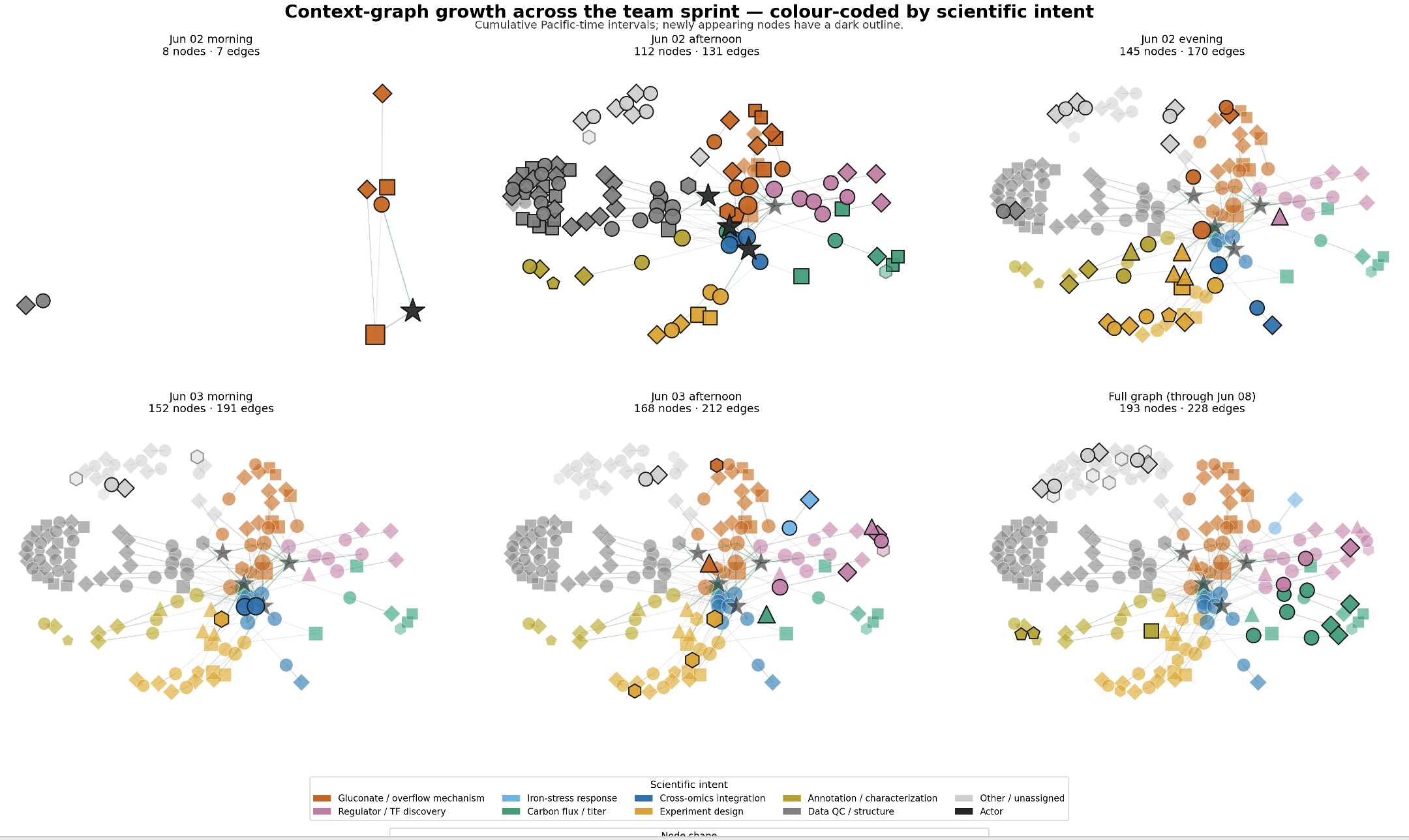}
\caption{\textbf{Graph evolution reveals the diversity of scientific intent.} Cumulative snapshots show isolated analyses converging into a unified model, with colors marking intents from data quality control (gray) and gluconate-overflow mechanisms (orange) to experiment design (yellow).}
\label{fig:graph-evolution}
\end{figure}

\textbf{Experiment recommendations grounded in distributed evidence:} The shared model informed an experiment plan authored by User-J. Fig.~\ref{fig:convergence} traces the three expert threads through the converged model (T2) into the final proposal (T3). The proposal mapped each model element to an intervention: \emph{mqo-2} perturbation/CRISPRi (overflow node), \emph{kguT} knockout/overexpression (import/re-assimilation), PP\_2256 deletion or sulfonate supplementation (stress context), and PP\_1794 characterization. Fig.~\ref{fig:graph-evolution} visualizes the team effort as the active context graph evolves from isolated analyses toward integrated experimental design.

\subsection{Impact of network scaling}
\label{sec:solo-baselines}

\textbf{Comparing network scaling with model scaling:} To isolate the effect of networked intelligence, we benchmarked the Mycelium-supported team sprint against two single agent autonomous baselines. The objective was to determine whether a standalone model, given identical data, maximum reasoning effort, and unconstrained autonomy could subsume the function of a distributed human-AI network. All executions used Claude Opus 4.8. Baseline B  was explicitly prompted with the multidisciplinary team specification, and baseline C was instructed to reason solely based on goal and data specification, without team specification (see Box S1 in Supplementary Method for full prompt). The single agent baselines were configured for autonomous operation with maximum reasoning effort (\texttt{xhigh}) and adaptive thinking \cite{anthropic_opus48_prompting}; Mycelium was run with the default \texttt{high} setting for inference cost and consistency. Both baselines ran to natural termination without hitting computational limits and successfully produced mechanistically grounded findings. 

\begin{table}[h]
\centering
\caption{\textbf{Networked execution expands scientific breadth over standalone baselines.} Evaluation of 26 trace-neutral artifacts scored from 0 (absent) to 4 (actionable). Networked execution maximizes global coverage without sacrificing the specificity of successfully surfaced artifacts.}
\label{tab:solo-product-summary}
\begin{tabularx}{\linewidth}{>{\raggedright\arraybackslash}Xccc}
\toprule
Metric & Mycelium & Agent B & Agent C \\
\midrule
Artifacts surfaced (Score $>0$) & \textbf{25} & 17 & 18 \\
Evidence-grounded or higher (Score $\geq 2$) & \textbf{22} & 14 & 15 \\
Experiment-ready or higher (Score $\geq 3$) & \textbf{17} & 9 & 11 \\
Actionable decision rule (Score $=4$) & \textbf{4} & 2 & 3 \\
Breadth-weighted artifact score (absences = 0) & \textbf{2.62} & 1.62 & 1.81 \\
Average specificity of surfaced artifacts & \textbf{2.72} & 2.47 & 2.61 \\
\bottomrule
\end{tabularx}
\end{table}


\textbf{Auditing the evidence-to-action pipeline:} We evaluated these executions by extracting a global set of 26 unique ``scientific artifacts'', defined as any traceable analytical claim, mechanistic hypothesis, or actionable decision rule capable of directing the execution of experimental protocols. Each execution trace was scored using an integer scale ranging from 0 (absent) to 4 (actionable, data-backed rationale) (Table~\ref{tab:solo-product-summary}; full matrix in Table \ref{tab:full-product-matrix}; mathematical formulations in Supplementary Method S2). This rubric captures the evidence-to-action pipeline rather than final-answer similarity alone \cite{chen2025scienceagentbench, mitchener2025bixbench, gao2026graph, wang2026agent}. It deliberately separates two metrics: \emph{breadth} (the fraction of the 26 artifacts explicitly surfaced, scoring $>0$) and \emph{specificity} (the level of actionability on the 1--4 scale that a surfaced artifact achieved). Because expert teams and autonomous models may possess tacit knowledge they do not explicitly articulate, this matrix is strictly an audit of explicit evidence-to-action coverage rather than a measure of latent capability.

\textbf{Networked execution expands discovery breadth:} In this evaluation, networked execution increased artifact breadth while preserving comparable specificity among the artifacts it surfaced. Mycelium generated 25 artifacts, including 22 evidence-grounded or higher and 17 experiment-ready, compared with 17 present and 9 experiment-ready for baseline B and 18 present and 11 experiment-ready for baseline C. As a breadth-weighted summary, the mean score over the full 26-artifact set, with absences scored as zero, was higher for Mycelium (2.62) than for baseline B (1.62) or baseline C (1.81). Unlike convergent tasks with definitive completion states (e.g. software engineering benchmarks \cite{jimenez2024swe}), open-ended scientific discovery relies on systemic exploration and continuous feedback to expose and correct analytical blind spots.

\textbf{A complementary scaling axis:} When the standalone baselines successfully surfaced an artifact, these models developed it with a specificity comparable to that of the networked team (average specificity: 2.72 for Mycelium vs. 2.47 and 2.61 for the baselines). This result indicates that the single-context baselines were capable of developing localized artifacts; in this evaluation, the primary difference was exploration breadth during open-ended discovery. Baseline C, which received no team-composition framing, outperformed baseline B, which was explicitly instructed to integrate findings as a multidisciplinary team would, in both artifact breadth and specificity. 

\begin{table}[h]
\centering
\footnotesize 
\setlength{\tabcolsep}{4pt} 
\renewcommand{\arraystretch}{1.1} 

\caption{\textbf{Network scaling expands discovery breadth over single-agent autonomous models.} The detailed scores show two trends. First, single-agent baselines are capable reasoners when localized on a specific topic. The autonomous baselines (B and C) generated 9 and 11 highly developed artifacts (scoring $\geq 3$), respectively. Notably, baseline C — which reasoned solely from the goal statement and data specifications without any team-composition framing, outperformed baseline B. By contrast, Mycelium's defining advantage is systemic breadth. It surfaced 25 of the 26 mapped artifacts, with 17 scoring 3 or higher. In this evaluation, networked human-AI execution provided a complementary axis of broader problem-space exploration alongside localized model reasoning.\\}
\label{tab:full-product-matrix}

\arrayrulecolor{black!20}

\begin{tabularx}{\linewidth}{>{\raggedright\arraybackslash}p{0.5cm} X >{\raggedright\arraybackslash}p{3.5cm} c c c}
\toprule
\rowcolor{black!6} 
\textbf{\color{black}ID} & \textbf{\color{black}Scientific Artifact} & \textbf{\color{black}Technical Category} & \textbf{\color{black}A} & \textbf{\color{black}B} & \textbf{\color{black}C} \\
\midrule
A01 & Nitrogen dominance over organic-acid and gluconate output & Bioprocess Engineering & \scF & \scF & \scF \\ \hline
A02 & Quantitative iron impact on organic-acid yield and saturation & Bioprocess Engineering & \scT & \scT & \scH \\ \hline
A03 & Iron gating mechanism of gluconate-to-organic-acid flux & Metabolic Mechanisms & \scH & \scH & \scF \\ \hline
A04 & Primary gluconate carbon routing via 2-ketogluconate branch & Metabolic Mechanisms & \scH & \scH & \scH \\ \hline
A05 & Ranked validation design for N $\times$ Fe with directional controls & Exp. Design \& QC & \scH & \scF & \scF \\ \hline
A06 & Constitutive dCas12a strain for slow-uptake/succinate platform & Strain Engineering & \scT & \scH & \scH \\ \hline
A07 & Landing-pad lineage optimization for glucose-to-gluconate conversion & Strain Engineering & \scT & \scH & \scH \\ \hline
A08 & \textit{mqo-2} malate-node CRISPRi target and promoter verification & Strain Engineering & \scF & \scT & \scZ \\ \hline
A09 & \textit{PtxS} CRISPRi targeting for the 2-ketogluconate carbon sink & Strain Engineering & \scH & \scZ & \scH \\ \hline
A10 & CRISPRi chassis and perturbation selection for validation & Strain Engineering & \scT & \scZ & \scH \\ \hline
A11 & Discovery of unannotated producer-associated protein PP\_1794 & Strain Engineering & \scF & \scZ & \scZ \\ \hline
A12 & HPLC–proteomics discordance in gluconate accumulation & Metabolic Mechanisms & \scF & \scO & \scO \\ \hline
A13 & Lack of canonical gluconate/ED operon induction at 12 hours & Metabolic Mechanisms & \scH & \scO & \scO \\ \hline
A14 & Identification of formate as a low-nitrogen metabolic byproduct & Analytical Chemistry & \scH & \scH & \scH \\ \hline
A15 & Integration of 2-ketogluconate as a targeted HPLC analyte & Analytical Chemistry & \scO & \scH & \scH \\ \hline
A16 & Anaerobic-AMP2 vs. obligate-aerobe platform compatibility risk & Bioprocess Engineering & \scZ & \scH & \scZ \\ \hline
A17 & Proteomics quality control, PCA packages, and imputation rules & Exp. Design \& QC & \scH & \scO & \scO \\ \hline
A18 & Genome-wide differential abundance and GO-enrichment packages & Exp. Design \& QC & \scH & \scT & \scT \\ \hline
A19 & \textit{GnuR} regulatory identity correction and specificity checks & Exp. Design \& QC & \scH & \scZ & \scZ \\ \hline
A20 & Sulfonate supplementation assays to modulate gluconate yields & Strain Engineering & \scH & \scZ & \scZ \\ \hline
A21 & Targeted \textit{kguT} knockout or overexpression strategies & Strain Engineering & \scH & \scZ & \scZ \\ \hline
A22 & Characterization of PP\_2256/\textit{dctA}-III regulatory island & Strain Engineering & \scH & \scZ & \scZ \\ \hline
A23 & Inoculum nitrogen carry-over mitigation and experimental control & Exp. Design \& QC & \scH & \scZ & \scT \\ \hline
A24 & Baseline evaluation of headroom for future yield optimization & Bioprocess Engineering & \scO & \scT & \scT \\ \hline
A25 & Regulatory hypothesis for \textit{FnrB} redox malate-node governor & Metabolic Mechanisms & \scT & \scZ & \scZ \\ \hline
A26 & Unit reconciliation for FeCl$_3$ across datasets and text & Exp. Design \& QC & \scO & \scT & \scT \\ 
\bottomrule
\end{tabularx}
\end{table}

\section{Discussion}

Networked intelligence accelerates discovery through active context propagation across users. In this case study, a shared active context graph enabled three domain experts to converge on a mechanistic model of carbon-overflow rewiring in \emph{P.\ putida} by generating a broader set of actionable scientific artifacts than autonomous agents operating over identical data (section \ref{sec:solo-baselines}). While this rapid acceleration compresses what is typically a four-to-six-month human-mediated iteration cycle into a single sprint, the primary breakthrough is architectural. Convergence emerged from a routed cross-modality link where a regulatory anomaly in one domain became a constraint on another (proteomic interpretation), which in turn formed the premise for physical experimental design.

\subsection{Calibrating agentic proactivity and claim attribution in human-AI networks}
\label{sec:Design variables}

Operating a distributed scientific context network introduces design parameters absent in standalone reasoning agents. The first is systemic proactivity \cite{zhang2024proagent, horvitz1999principles}, which requires calibrating the autonomous background investigation effort against user review burden. In this campaign, Mycelium operated primarily responsively, but a critical, sustained active fraction of self-initiated investigation (averaging roughly 18\% of total system actions, with peaks up to 37\%) drove discovery (Fig.~\ref{sfig:workload} in Supplementary Methods). This active fraction was necessary for cross-user context propagation, as the system proactively identified and routed critical constraints across independent participant threads without explicit user prompts. Proactivity is therefore a central calibration problem for scientific networks: deficient proactivity drops critical links, while excessive proactivity overwhelms human oversight.

The second parameter encompasses claim attribution, the process of tracking the precise human or agent origin of a scientific finding, and epistemic weighting, which determines the degree of analytical trust the system assigns to that finding based on its source \cite{wang2026agent, pearl2022fusion}. Shared evidence and its attached provenance flowed asymmetrically among all three researchers and the system actor (Fig.~\ref{sfig:workload}). The current architecture attributes every propagated claim to its source but applies equal epistemic weight. While equal attribution supports strict accountability \cite{wilkinson2016fair}, future architectures must model differential expertise \cite{jacobs1991adaptive}, preferring reliable contributors or surfacing unresolved contradictions rather than a silent merge.

\subsection{Network scaling overcomes the exploration bottlenecks of standalone agents}

A central event in this deployment was a change in the functional role of a scientific claim. A local observation reached another user in the network and altered the team’s analysis. Standalone agents miss cross-domain linking opportunities due to the lack of continuous, asynchronous interaction of independent perspectives. Here, networked intelligence denotes a capability in which humans, agents, and instruments operate in separate contexts that are connected by a runtime that routes relevant state when it is expected to affect downstream action. This capability complements model scaling. Stronger models and larger context windows can improve localized synthesis, but they do not by themselves resolve data rights, institutional firewalls, or operational knowledge held in separate contexts. Where these contexts cannot be merged into a single window, networked execution can provide a complementary scaling axis.

Computationally, the architecture performs sparse conditional computation \cite{jacobs1991adaptive,shazeer2017outrageously} over distributed contexts. over. Many claims may have limited cross-domain utility. The challenge is to identify the links that improve a receiving context’s expected decision value. We formalize this routing value in the Supplementary Note, clarifying the mathematical boundary between operations a scaled standalone agent can efficiently subsume and those that remain strictly irreducible due to non-mergeable contexts.

\subsection{A standardized protocol supports distributed reasoning state}

Networked intelligence requires a standardized contract for agents, instruments, and researchers to exchange structured scientific state. Independent actors adhere to a shared schema, writing typed, provenanced entries directly into the active context graph \cite{wilkinson2016fair,soilandreyes2022rocrate}. Consumers such as experimental planners \cite{mada2025}, safety monitors \cite{osprey2026}, literature agents \cite{ghareeb2026multi}, or human researchers can then query hypotheses and lineage as structured data rather than re-deriving them from free text. Components that honor this data model can contribute results with persistent attribution and receive routed context across sessions. The protocol may support applications that standalone tools do not readily provide, including routing physical experiments from a project’s global state and cross-project fusion \cite{academy2025} that links hypotheses across laboratories subject to privacy constraints.

\subsection{Limitations and outlook}

The present deployment provides evidence that active context propagation can alter scientific interpretation and experimental action. Next priority is scaling this runtime to validate reliability across multi-laboratory consortiums and extended, multi-month investigations. Beyond systemic scale, resolving conflicting signals across routed contexts (Section~\ref{sec:Design variables}) requires sophisticated handling. Finally, sharing context across distinct scientific disciplines introduces translation risks. While Mycelium tracks the exact origin of a claim, future systems must also measure the uncertainty and loss of specific meaning that occurs when a finding is adapted from one domain to another.



\section{Acknowledgements}

This work was supported by the OPAL project, supported by the U.S. Department of Energy (DOE), Office of Science, Office of Biological and Environmental Research (BER) and Office of Advanced Scientific Computing Research (ASCR). A portion of this research was performed on a project award at the Environmental Molecular Sciences Laboratory (EMSL), a DOE Office of Science User Facility sponsored by the BER program, using the Anaerobic Microbial Phenotyping Platform (AMP2), under Contract No. DE-AC05-76RL01830. A portion of this research was supported by the ``Accelerating biological discovery by enabling genomic resources in Deep Phenotyping on automated platforms" project, funded by DOE BER and ASCR, under FWP 86450. Pacific Northwest National Laboratory is operated by Battelle for the U.S. DOE under Contract No. DE-AC05-76RL01830.

\section{Author Contributions}

S.C. conceived networked intelligence, designed and implemented the Mycelium system, and wrote the manuscript. J.C., L.M., and E.B. conducted the multi-omics campaign through Mycelium and provided scientific interpretation of the routed findings. A.Be. provided the engineered strains developed under his BER biodesign program and contributed to strain design. J.E. constructed the engineered strains. P.P. designed the proteomics experiments. K.T. prepared samples. S.B. provided access to and oversight of the AMP2 platform at EMSL. Y.G. and A.Bi. performed proteomics data analysis. K.S. performed statistical analysis. K.B.J., R.R., C.O., J.B., and K.W. contributed to study conceptualization and scientific framing, advised on experimental design and feasibility, and provided supervision and resources. All authors reviewed and approved the manuscript.

\clearpage
\section*{Supplementary Note: Networked vs Monolithic Intelligence}\addcontentsline{toc}{section}{Supplementary Note: Networked intelligence as sparse conditional computation}
\newtheorem{definition}{Definition}
\newtheorem{claim}{Claim}
\newtheorem{corollary}{Corollary}

This subsection formulates human–AI team science as sparse conditional computation over distributed contexts. It considers three questions:

\begin{enumerate}
    \item[\textbf{Q1.}] Under what conditions can a monolithic AI model with large context windows and broad data access perform the same function as a collaborative network?
    \item[\textbf{Q2.}] Under what conditions does a networked approach retain an advantage that cannot be recovered by increasing the context window of a single model?
    \item[\textbf{Q3.}] When does a human expert contribute value beyond that of an agent operating on comparable information?
\end{enumerate}

The discussion uses the empirical campaign in Section 3 as a running example. In that campaign, Mycelium routed findings among three researchers working in separate contexts. The example illustrates the proposed mechanism, rather than establishing a universal separation between networked and monolithic systems.

\textbf{Architectural Premise:} Networked intelligence can implement sparse conditional computation over distributed human–AI contexts. The architecture partitions the scientific workload across reasoners, scientists, agents, and instruments. Only a subset of local findings is likely to be useful outside the context in which it arose. The active context graph $G=(V,E)$ maintains the shared state and supports the identification of candidate cross-context links.

The central routing question is whether propagating a finding to a neighboring context improves that context’s downstream experimental decisions.

\textbf{Context and Routing Value:} We define a context $C_i=(X_i,H_i,A_i)$ as a localized workspace containing local evidence $X_i$, an active hypothesis state $H_i$, and a set of actionable experimental interventions $A_i$.

\textbf{Example (The Context):} Consider the domain expert User-J from the multi-omics campaign. User-J's context $C_J$ contains local high-performance liquid chromatography (HPLC) measurements ($X_J$), an active hypothesis regarding carbon-sink rewiring ($H_J$), and a set of physical interventions that can be authorized on the laboratory platform, such as a kguT knockout or mqo-2 perturbation ($A_J$). 

The decision value of a context under hypothesis state $H$ is the maximum expected utility of an action:
$$V_i(H)=\max_{a\in A_i}\mathbb{E}_{\omega\mid H}\,[\,U_i(a,\omega)\,]$$ When a different context produces a new graph node $v$ (such as an observation or finding), the routing value of $v$ to context $j$ is:$$\Delta(v\to j)=V_j(H_j\oplus v)-V_j(H_j)$$

where $H_j\oplus v$ represents the receiver's localized belief state updated via a provenance edge (e.g., \texttt{derived\_from}). A candidate edge is a \emph{critical edge} when $\Delta(v\to j)>\tau$, where $\tau$ for a system-defined threshold. Estimating this quantity requires an implementation-specific model of belief updates, expected utility, and recipient attention costs. Under this formulation, the system routes a finding only when its estimated decision value exceeds the threshold.

\textbf{Example (The Routing Value):}  User-E identified a regulatory anomaly in an isolated context: canonical gluconate-uptake genes failed to express as expected. Mycelium scores and routes this finding to User-L and User-J. Integrating this new claim fundamentally alters their working hypotheses ($H_J \oplus v$), shifting their analysis from a pathway-induction problem to an overflow-rewiring problem. If that update changes the intervention selected from ($A_J$) and improves its expected utility, then the routing value ($\Delta$) is positive.

\textbf{The Dilemma:} With routing value formalized, we must address the architectural critique: \textit{Why build a distributed network to find these rare connections, rather than simply passing all the data into a single, massively scaled AI model?}

The answer splits into two regimes. The first regime directly answers \textbf{Q1}, outlining where a monolithic model subsumes the network through mere efficiency.

\textbf{Claim 1 [Efficiency Regime]:}
Suppose a project generates $N$ claims across $K$ expert contexts. A centralized model could ingest each of the $N$ claims once. Its cost depends on its architecture and the length or structure of its input, not automatically on $NK$. Conversely, the routing system must discover the relevant $d$ recipients. If recipient discovery requires comparing each claim with every context, it can itself cost $O(NK)$. Thus, $O(Nd)$ follows only with an additional assumption that relevant recipients can be found cheaply.

This efficiency gain holds only when recipient discovery can be performed at lower cost than dense comparison and when (d) is substantially smaller than (K). This does not establish a rigorous separation between networked and monolithic intelligence. A standalone AI agent with a massive context window could theoretically ingest the entire proteomics table, the HPLC readings, and the transcription factor library. It could perform the sparse arithmetic internally, ignore the irrelevant data, and discover the exact same pathway overlap. Therefore, computational efficiency alone cannot justify a networked architecture.

To answer \textbf{Q2} and prove the network is strictly necessary, we must enter the second regime. This regime defines the exact boundaries that a single AI model cannot cross, regardless of its size or context length. Here, the factored network performs a fundamentally irreducible computation.

\textbf{Claim 2 [Irreducibility Regime]:}
A factored network suceeds whenever either of two conditions holds:
\emph{(i)~Independent corroboration:} Reducing error via agreement requires estimators with conditionally independent error profiles. Because a single monolithic model conditions its outputs on a shared parameter distribution, its errors across domains remain correlated. Robust epistemic corroboration requires integrating truly independent priors, which are provided by a diverse network of distinct human experts and specialized agents or tools.
\emph{(ii)~Non-mergeable contexts:} Real-world team science frequently involves evidence that cannot physically or organizationally be placed into a single context window.

Condition (i) provides a formal account for why a network must preserve distinct expert models. A single AI model uses a single set of weights. If it encounters an edge case and makes a confident error early in the analysis, it will mathematically propagate that exact same error across every subsequent step of the workflow. An independent expert acts as an epistemic circuit breaker: their value is that their failure modes are entirely different from the primary AI's failure modes.

Condition (ii) explicitly defines why the monolithic AI system will never have full knowledge of the problem. A super-human AI can only be super-human if it has access to all the variables. In scientific research, the full state of the project is never fully digitized or perfectly observable. A standalone AI cannot factor in the human's tacit knowledge about a scientific instrument, the proprietary regulatory policies of a partner institution, or the physical constraints of the laboratory floor. Because this local evidence is held in the human expert or behind a physical firewall, dense model integration is impossible.

\textbf{Answering Q3: When does a Human Expert outperform an Agent?} 
Let an expert thread $i$ map local evidence $X_i$ to a context-specific posterior distribution $q_i(h\mid X_i)$ over hypotheses, and let $q_0$ represent a generic standalone reasoner. The expert provides strict systemic value when, for a productive mechanistic hypothesis $h^\star$, $q_i(h^\star)>q_0(h^\star)$. 

In plain terms, $h^\star$ is the hypothesis with highest utility (e.g., ``The bug is in incorrect PCA computation," or ``The gluconate is accumulating due to overflow rewiring"). $q_i(h^\star)$ is the probability that the expert assigns to that correct hypothesis, given the data they can see ($X_i$). $q_0(h^\star)$ is the probability that a generic, standalone AI model assigns to that correct hypothesis. Therefore, $q_i(h^\star)>q_0(h^\star)$ simply means the expert is sharper than the generic model. The expert places more probability mass on the correct answer.

This corollary explicitly defines the mathematical threshold where an expert becomes obsolete versus where they remain essential. If a task is straightforward and fully observable such as generating standard workflow software implementations and the AI model demonstrably identifies the optimal solution ($h^\star$) with super-human empirical performance, adding a human expert adds no accuracy to the baseline prediction.

\textbf{Conclusion:} The value of a human-AI network does not rest on a general claim that human cognition is more expressive than AI models. Its value arises when high-quality evidence, action authority, or independent judgment remains distributed across contexts. Networked intelligence is therefore best understood as a conditional systems capability. It can improve scientific coordination when relevant information cannot be fully merged, when sparse routing is less costly than dense integration, and when the network preserves useful diversity in evidence and error profiles.

\clearpage
\section*{Supplementary Methods}
\addcontentsline{toc}{section}{Supplementary Methods}
\renewcommand{\thefigure}{S\arabic{figure}}
\setcounter{figure}{0}

\subsection*{S1. Standalone agent baseline: harness, sweep, and prompts}

The team-framed prompt is reproduced in Box~S1, lightly abridged to omit field-level output schemas. Both variants receive the same frozen problem statement, datasets, and annotation and pathway resources, and the same artifact contract; the unframed variant differs only in the highlighted clause.
 
\definecolor{promptframe}{HTML}{2F6F4E}
\definecolor{sweptclause}{HTML}{D9ECDF}
\definecolor{phgray}{HTML}{8A8A8A}
\newcommand{\ph}[1]{{\color{phgray}\ttfamily\{#1\}}}
\sethlcolor{sweptclause}
 
\begin{tcolorbox}[
  colback=white, colframe=promptframe, boxrule=0.6pt, arc=2pt,
  left=8pt, right=8pt, top=6pt, bottom=6pt,
  title={\textbf{Box~S1\,$\vert$\, Standalone agent baseline prompt (team-framed variant)}},
  fonttitle=\small, fontupper=\small\setlength{\parskip}{4pt}]
 
\textbf{Role and objective.}~You are a senior microbial systems biologist analyzing a multi-omics dataset from \emph{Pseudomonas putida} KT2440, working alone. \hl{Cover every relevant angle yourself and integrate across them as a multidisciplinary team would.} Using proteomics and HPLC data, predict molecular insights into how modification of gluconate and organic-acid production, including environmental and genetic factors in the culture medium, affects phenotype and the production of compounds of interest. Design up to ten validation experiments to be run on the Anaerobic Microbial Phenotyping Platform (AMP2).
 
\ph{workspace\_description}
 
\textbf{Environment.}~Run any analysis you need in the provided code-execution environment, including your own scripts: statistics, dimensionality reduction, differential analysis, enrichment, correlation, clustering, and arbitrary Python over the provided files and annotation tables.
 
\textbf{Tagging.}~Tag meaningful intermediate findings: \texttt{\#idea} (a claim or hypothesis; fields: statement, tests), \texttt{\#question} (an unmet need; fields: need, success criterion), \texttt{\#result} (a synthesized finding; fields: statement, driver, computation), \texttt{\#note} (an observation worth keeping; fields: statement, context).
 
\textbf{Deliverable.}~Before the run ends, produce and save (1)~a ranked list of up to ten validation experiments over nitrogen (\texttt{NH4\_2SO4\_g\_per\_L}) $\times$ iron (\texttt{FeCl3\_ug\_per\_L}) combinations with predicted outcomes, (2)~a summary of the key reasoning chain naming which data patterns drove the predictions, and (3)~a log of all tagged entries.
 
\textbf{Execution.}~Work at maximum reasoning effort in a single continuous context, autonomously to completion. For minor choices, make a reasonable assumption and note it rather than stopping to ask.
 
\textbf{Final artifact contract.}~End the final answer with two machine-readable fenced blocks, \texttt{experiment\_design.csv} (one ranked experiment per row) and \texttt{reasoning\_chain.jsonl} (one tagged entry per line). \ph{column and field semantics omitted; see released harness}
\end{tcolorbox}
 
\noindent\footnotesize The \hl{highlighted} clause is the swept framing. The unframed variant removes it, so the model is still told it is working alone but is not told to integrate as a multidisciplinary team would; every other instruction, the data, and the artifact contract are identical across the two.\normalsize

\subsubsection*{S2. Evaluation Metrics Formulation}
To formalize our evaluation of the evidence-to-action pipeline, we define the global set of 26 mapped scientific artifacts as $A$. For a given execution trace, the assigned score for an artifact $i \in A$ is denoted as $S_i$, where $S_i \in \{0, 1, 2, 3, 4\}$.

We calculate the \textbf{breadth-weighted artifact score} as the arithmetic mean over the entire artifact set, penalizing absences (where $S_i = 0$):
$$ \text{Breadth-Weighted Score} = \frac{1}{|A|} \sum_{i \in A} S_i $$

To isolate the depth of reasoning for successfully identified artifacts, we calculate the \textbf{average specificity}. This is the mean score of only the subset of explicitly surfaced artifacts, $A_{\text{surfaced}}$, where $S_i > 0$:
$$ \text{Average Specificity} = \frac{1}{|A_{\text{surfaced}}|} \sum_{i \in A_{\text{surfaced}}} S_i $$

\begin{figure}[!h]\centering\includegraphics[width=0.8\linewidth]{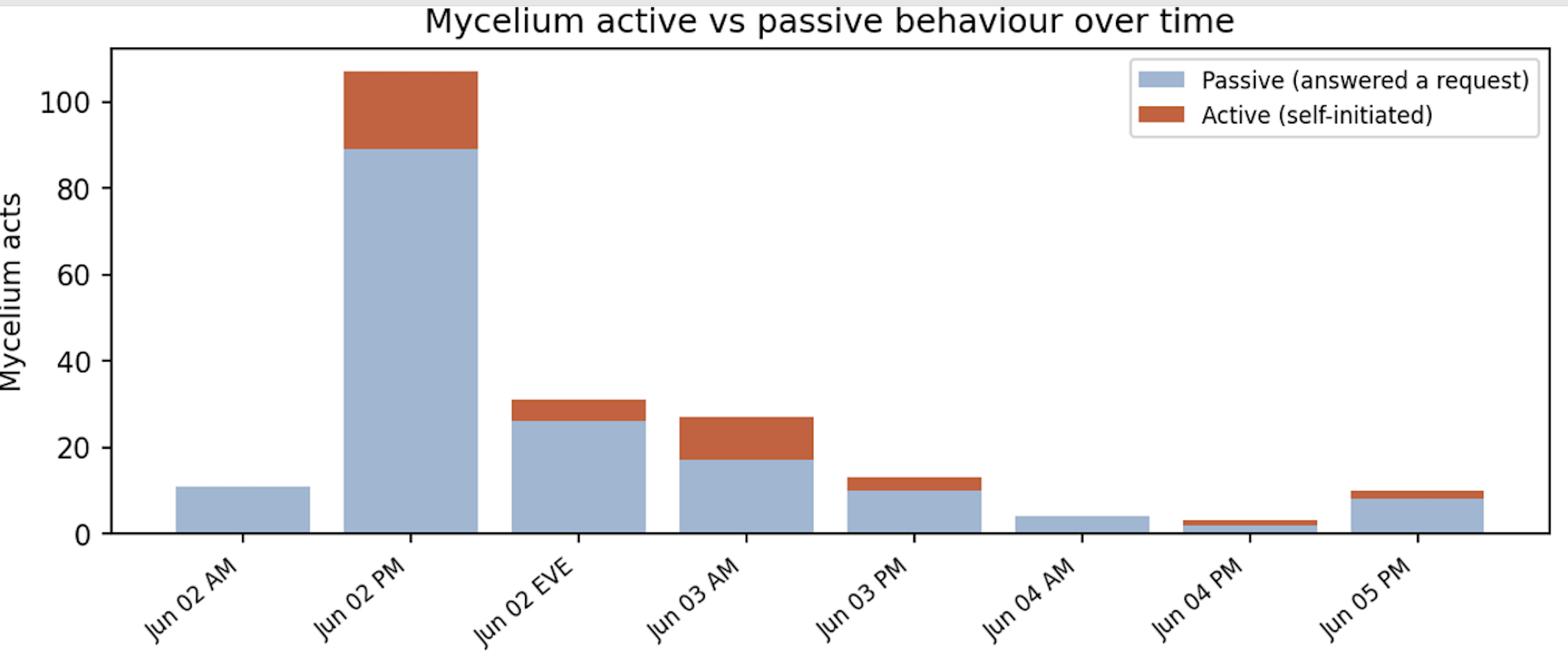}\caption{\textbf{Workload distribution, analytical-intent diversity, and cross-actor propagation.} Mycelium activity was predominantly responsive to participant requests, with a smaller sustained fraction of active background investigation, context routing, and system-authored findings. }\label{sfig:workload}\end{figure}

\begin{figure}[h]
\centering
\includegraphics[width=1\linewidth]{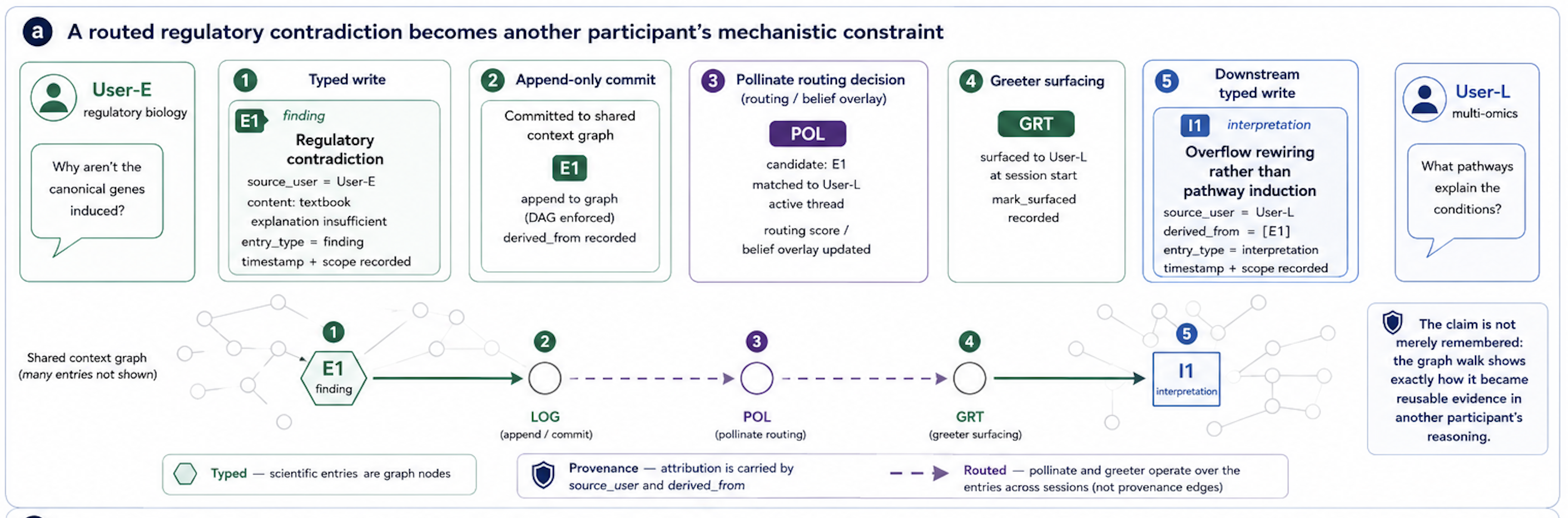}
\caption{\textbf{Illustration of cross-context propagation and provenance tracking.} This figure shows an trace from the multi-user study reported in the results section.  A regulatory contradiction identified by User-E is captured as shared scientific state, automatically routed to User-L, and incorporated as a constraint on the downstream mechanistic model. The numbered sequence shows how Mycelium records, routes, surfaces, and reuses the finding while preserving its attribution and provenance from the original entry, E1, to the resulting interpretation, I1.}
\label{fig:propagation}
\end{figure}

\clearpage

\section*{Supplementary Dataset}
\addcontentsline{toc}{section}{Supplementary Dataset}

The experimental campaign comprises a four-modality dataset structured to address a cross-modal reconciliation problem. Five individual data tracks span proteomics, a curated metabolic network, extracellular metabolite chemistry, and media-design phenotypes (Table~\ref{stab:dataset}). Each dataset resolves a distinct facet of the system's phenotype: identifying protein variance, mapping pathway context, verifying secreted chemistry, or isolating key media drivers.

Table~\ref{stab:dataset} details these datasets, their dimensions, design factors, and corresponding scientific roles. The underlying biological evaluation spans four distinct strains of \emph{P. putida}: wild-type KT2440, a landing-pad (LP) lineage, and two LP-background dCas12a variants (constitutive and induced). This dCas12a contrast establishes the mechanistic baseline for the gluconate-retention findings detailed in the main text.

\begin{table}[h]
\centering
\footnotesize
\setlength{\tabcolsep}{5pt} 
\renewcommand{\arraystretch}{1.3} 

\caption{\textbf{Multi-modal dataset architecture, experimental design factors, and operational scopes within the Mycelium-driven \emph{P. putida} campaign}. Rows define the structural scale, factor levels, and analytical objectives (Scientific role) required for cross-modal integration. Media-optimization mass concentrations utilize distinct scales by design: g/L for nitrogen components and \textmu g/L for iron components. Quantitative combinatorial boundaries of the resulting hypothesis space are itemized in the design summary below.}
\label{stab:dataset}
\vspace{1cm}
\arrayrulecolor{black!15}

\begin{tabularx}{\linewidth}{@{} >{\raggedright\arraybackslash}p{2.4cm} >{\raggedright\arraybackslash}p{1.8cm} >{\raggedright\arraybackslash}p{2.2cm} X >{\raggedright\arraybackslash}p{3.0cm} @{}}
\toprule
\rowcolor{black!6} 
\textbf{Dataset} & \textbf{Modality} & \textbf{Size} & \textbf{Design factors (levels)} & \textbf{Scientific role} \\
\midrule
Proteomics (differential abundance) & Proteomics & 4{,}495 proteins, 62 samples & 4 strains $\times$ 2 carbon $\times$ 2 timepoints (16 conditions, up to four replicates) & Which proteins change (User-L) \\ \hline
UniProt annotation & \cite{uniprot2025uniprot} & 5{,}564 entries $\times$ 13 fields & KEGG, EC, GO (BP/CC/MF), pathway & Functional grounding for enrichment \\ \hline
Metabolic pathway graph & Curated network from \cite{kanehisa2025kegg} & 1{,}170 nodes / 1{,}007 edges & KEGG-reaction-labeled edges & Pathway context for proteomic signals \\ \hline
HPLC phenotype & Metabolite quantitation & 48 measurements, 4 analytes & 4 strains $\times$ 2 carbon $\times$ 3 timepoints (24 conditions) & What the secreted chemistry confirms (User-J) \\ \hline
Media-optimization DoE & Design / phenotype & 75 measured + 28 proposed runs & continuous (NH$_4$)$_2$SO$_4$ $\times$ FeCl$_3$ & The actionable intervention axis on AMP2 \cite{smith2026using} \\ \hline
TF library & Reference & 392 regulators $\times$ 10 fields & gene-identifier keyed & Regulatory grounding (User-E) \\ 
\midrule
\rowcolor{white} 
\multicolumn{5}{@{}p{\linewidth}@{}}{\textit{Design summary.}~16 proteomic and 24 phenotype conditions admit 120 and 276 pairwise contrasts as the hypothesis space a mechanism must be consistent with. The validation action space is continuous, not a fixed grid: 28 distinct nitrogen and 28 distinct iron levels, no repeated point. Taking the 28 proposed media points as the candidate set gives $28 \times 2~\text{carbon} \times 4~\text{strains} = 224$ candidate experiments, which the team must reduce to ten.} \\
\bottomrule
\end{tabularx}
\end{table}


\clearpage

\bibliographystyle{ieeetr}
\bibliography{refs}
\end{document}